\documentclass[letterpaper, 10 pt, conference]{ieeeconf}  % Comment this line out if you need a4paper

\IEEEoverridecommandlockouts                              % This command is only needed if 
\usepackage{times}
\usepackage{epsfig}
\usepackage{graphicx}
\usepackage{amsmath}
\usepackage{amssymb}
\usepackage{caption}
\usepackage{subcaption}

\title{\LARGE \bf
Dense Piecewise Planar RGB-D SLAM for Indoor Environments
}

\author{Phi-Hung Le and Jana Kosecka% <-this % stops a space
\thanks{Department of Computer Science, George Mason University, 4400 University Drive MSN 4A5, Fairfax, Virginia 22030, USA
       {\tt\small \{ple13,kosecka\}@gmu.edu}}%
}

\begin{document}
\maketitle
\thispagestyle{empty}
\pagestyle{empty}

%%%%%%%%%%%%%%%%%%%%%%%%%%%%%%%%%%%%%%%%%%%%%%%%%%%%%%%%%%%%%%%%%%%%%%%%%%%%%%%%
\begin{abstract}

The paper exploits weak Manhattan constraints to parse the structure of indoor environments from RGB-D video sequences in an online setting.  We extend the previous approach for single view parsing of indoor scenes to video sequences and formulate the problem of recovering the floor plan of the environment as an optimal labeling problem solved using dynamic programming. The temporal continuity is enforced in a recursive setting, where labeling from previous frames is used as a prior term in the objective function. In addition to recovery of piecewise planar weak Manhattan structure of the extended environment, the orthogonality constraints are also exploited by visual odometry and pose graph optimization. This yields reliable estimates in the presence of large motions and absence of distinctive features to track. We evaluate our method on several challenging indoors sequences demonstrating accurate SLAM and dense mapping of low texture environments. On existing TUM benchmark~\cite{sturm11rss-rgbd} we achieve competitive results with the alternative approaches which fail in our environments.  

\end{abstract}

%%%%%%%%%%%%%%%%%%%%%%%%%%%%%%%%%%%%%%%%%%%%%%%%%%%%%%%%%%%%%%%%%%%%%%%%%%%%%%%%
\section{Introduction}

The paper exploits weak Manhattan constraints~\cite{SaurerVICOMOR12} to parse the structure of indoor environments from RGB-D video sequences. Manhattan constraints assume that all the planar structures in the environment are aligned with one of the axes of a single  orthogonal coordinate frame. In our setting the structure of the scene is comprised of sets of vertical  planes, perpendicular to the floor and grouped to different Manhattan coordinate frames. 
The problem of geometric scene parsing, in our case the floor plan recovery, involves more than just plane fitting and identification of wall and floor surfaces, which are supported by depth measurements. It requires reasoning where the walls intersect, what their extent is and what the occlusion boundaries are, especially in the case of missing or ambiguous depth measurements. Previous researchers studied the problem of scene parsing in the presence of Manhattan constraints in a single view setting. Several works tried to infer the scene structure using vanishing points and lines~\cite{UrtasunIndoorCVPR12} and alternative volumetric constraints~\cite{Lee-CVPR09}, using RGB images only~\cite{popupslam}.

\begin{figure}[htb]
\centering
\includegraphics[width=0.2\textwidth]{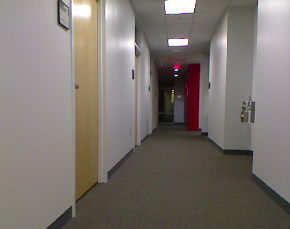}
\hspace{0.05cm}
\includegraphics[width=0.2\textwidth]{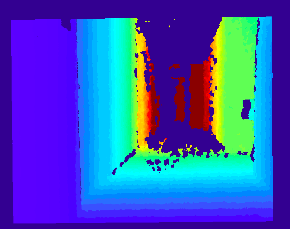} \\
\vspace{0.2cm}
\includegraphics[width=0.2\textwidth]{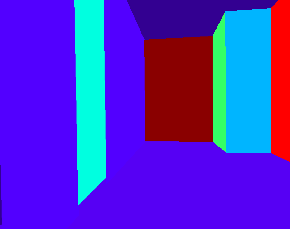}
\hspace{0.05cm}
\includegraphics[width=0.2\textwidth]{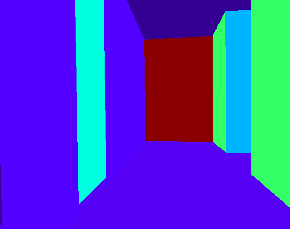}
\caption{Top: RGB image and the depth image; Bottom Single view parsing and temporal parsing results. Note the correctly parsed geometric structures in the right results. The colors correspond to different plane labels.}
\end{figure}
We adopt an approach for single view parsing from RGB-D views proposed in ~\cite{TaylorRSS12}. In this work the authors infer the 3D layout of the scenes from a single RGB-D view and pixel level labeling in terms of dominant planar structures aligned with the orientations determined by a Manhattan coordinate frame. The optimal labeling is carried out using dynamic programming over image intervals determined using geometric reasoning about presence of corners and occluding boundaries in the image. In our setting we relax the single Manhattan frame assumption and consider the set of dominant planes perpendicular to the floor, but at varying orientations with respect to each other. We further extend the approach to sequences and show how to formulate the geometric parsing recursively, by updating the single view energy function using previous parsing results. The proposed approach yields better, temporally consistent results in challenging RGB-D sequences.  In addition to the estimates of piecewise planar models, we use the Manhattan constraints for estimation of visual odometry in challenging sequences with low texture and large displacements and blur. The compact global models of indoor environments are then obtained by loop closure detection and final pose graph optimization~\cite{GraphSLAMTutorial} enabling globally consistent models. We  carry out extensive experiments to evaluate our approach. 

In summary, our contributions are:
\begin{itemize}
\item An extension of a geometric parsing approach for a single RGB-D frame to a temporal setting; 
\item An integration of structures inferred from the parsing step and point features to estimate accurate visual odometry, yielding drift free rotation estimates; 
\item These two components along with planar RGB-D SLAM, loop closure detection and pose graph optimization 
enable us to obtain detailed and high quality floor plan including non-dominant planar structures and doors.
\end{itemize}

\section{Related Work}

This work is related to the problem of 3D mapping and motion estimation of the camera from RGB-D sequences. This is a long standing problem, where several existing solutions are applicable to specific settings~\cite{KinectFusionISMAR11, NiesnerACM13, PollefeysIJCV08} . Many successful systems have been developed for table top settings or small scale environments at the level of individual rooms. These environments often have a lot of discriminative structures making the process of data association easier. The camera can often move freely enabling denser sampling of the views, making local matching and estimation of odometry well conditioned. Several approaches and systems have been proposed to tackle these environments and typically differ in the final representation of the 3D model, the means of local motion computation using either just RGB or RGB-D data and the presence or absence of the global alignment step.  

For the evaluation of visual odometry approaches only, Freiburg RGB-D benchmark datasets~\cite{sturm11rss-rgbd} are the de-facto standard. Simultaneous mapping and dense reconstruction of the environments has been successful in smaller workspaces, using a variety of 3D representations, including signed distance functions, meshes or voxel grids~\cite{KinectFusionISMAR11, KoltunCVPR15,NiesnerACM13}. Volumetric representations and an on-line pose recovery using higher quality LIDAR data along with final global refinement were recently proposed~\cite{WangCVPR16}, with more detailed related work discussion found within.  Approaches for outdoor 3D reconstruction and mapping of outdoors environments have been demonstrated in~\cite{PollefeysIJCV08}.

Another set of works focuses on the use of Manhattan constraints to improve 3D reconstruction either from a single view or multiple registered views as well as 3D structure. In~\cite{Flint-ICCV11} authors focused more detailed geometric parsing into floor, walls and ceiling using stereo, 3D and monocular cues using registered views. In~\cite{flint:etal:cvpr2010} the authors demonstrated an on-line real-time system for semantic parsing into floor and walls using a monocular camera, with the odometry estimated using a Kalman filter. The reconstructed models were locally of high quality, but of smaller extent considering only few frames. In~\cite{AlejoAR2015} the authors proposed a monocular SLAM framework for low-textured scenes and for the ones with low-parallax camera motions using scene priors. In~\cite{DPPTAMiros2015} a dense piecewise monocular planar SLAM framework was proposed. The authors detected planes as homogeneous-color regions segmented using superpixels and integrated them into a standard direct SLAM framework. Additional, purely geometric approaches assumed piecewise planarity~\cite{trevor12:planar_surfac_slam_sensor} and used multiple sensing modalities to reconstruct larger scale environments. The poses and planes were simultaneously globally refined using the final pose graph optimization. These works did not pursue more detailed inference about corners and occlusion boundaries induced by planar structures and estimated the planar structures only where the depth measurements were available. This is in contrast to pixel level labeling schemes of~\cite{Flint-ICCV11} where each pixel in the RGB frame is assigned a label. 
The more general problem of 3D structure recovery in indoors scenes has been tackled in~\cite{Furukawa-ICCV09} using denser high quality laser range data and a box like modeling fitting approach. This approach is computationally expensive and suitable for strictly Manhattan box worlds. 

Earlier works of~\cite{Lee-Hebert-Kanade-NIPS10} presented an approach for estimating room layour with walls aligned with Manhattan coordinate frame. An attempt to model the world as a mixture of Manhattan frames has been done in~\cite{straub2014mmf} where Manhattan mixtures were estimated in the post processing stage to refine the quality of the final model. In our case we handle this in an online setting. In~\cite{straub2015cvpr} the 3D rotation for visual odometry in an indoor Manhattan World is tracked by projecting directional data (normal vectors) on a hypersphere and by exploiting the sequential nature of the data. An effective approach for single RGB-D view parsing  was proposed in~\cite{TaylorRSS12}, where optimal plane labeling was obtained using a dynamic programming approach over a sequence of intervals in which were obtained by aligning the view with the gravity direction.  
 
The presented work extends single view parsing to video sequences. We show how to change the optimization to include the information from the previous frames and relax the Manhattan assumption, by considering vertical planes perpendicular to the floor. The relative orientation between the frames is estimated from consecutive single view estimates, requiring only single 3D point correspondence to estimate the relative translation between the views. 

\begin{figure}[htb]
\centering
\includegraphics[width=0.4\textwidth]{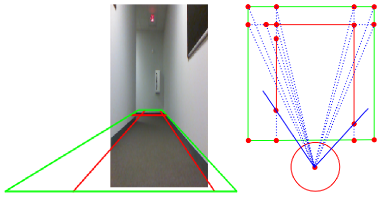}
\caption{Right: Bird's eye view of the line endpoints and intersections of all possible lines in the bounding box volume. Left: Intersections superimposed over image. Blue lines: the FOV of the camera. Red line segments: the projection of walls on the ground floor. Green line segments: the projection of the bounding box volume on the ground floor.}
\label{fig:intervals}
\end{figure}

Closest to our approach is the work in ~\cite{popupslam}. The authors developed a real-time monocular plane SLAM incorporating single view scene layout understanding for low texture structural environments. They also integrated planes with point-based SLAM to provide photometric odometry constraints as planar SLAM can be easily unconstrained. Our single view parsing attains higher quality of 3D models (including doors) and is tightly integrated with pose optimization and loop closure detection.

\section{Approach} 
\subsection{Single View Parsing}
\begin{figure}[htb]
\centering
\includegraphics[width=0.15\textwidth]{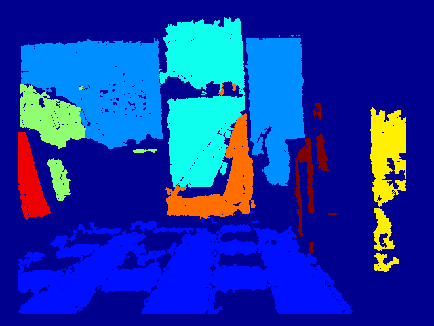}
\includegraphics[width=0.15\textwidth]{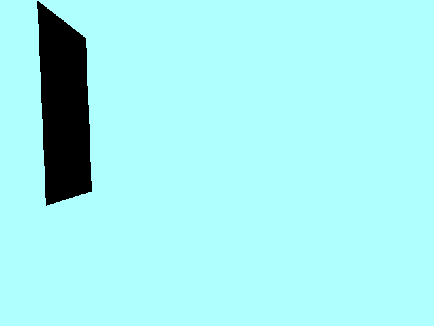}
\includegraphics[width=0.15\textwidth]{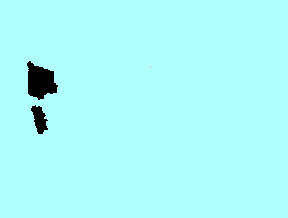}
\caption{Left: Best wall for each pixel. Middle: A projected quadrilateral from an interval to a wall. Right: support pixels for the wall being considered. }
\label{fig:supports}
\end{figure}

\begin{figure}[htb]
\centering
\includegraphics[width=0.5\textwidth]{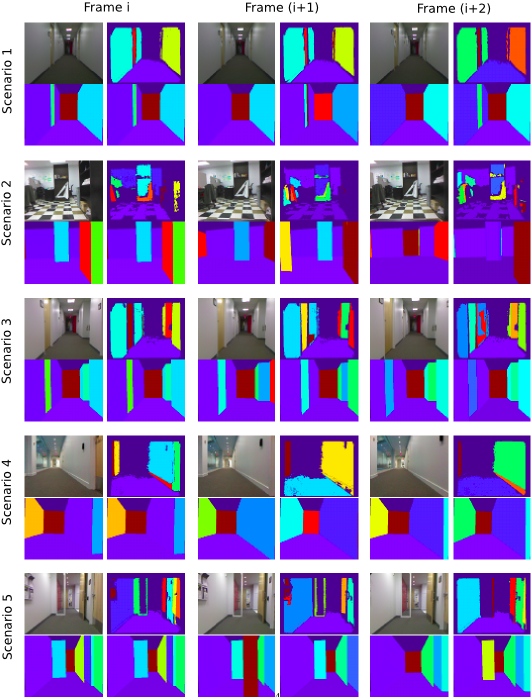}
\caption{Each row has three blocks showing the result of three consecutive frames. For each block, the top left is the RGB image, the top right are the walls that align with the dominant Manhattan frame, the bottom left is the result of single view parsing, and the bottom right is the result of temporal parsing.}
\label{fig:results}
\end{figure}

This paper extends the work of~\cite{TaylorRSS12} in which authors proposed a dynamic programming solution for single view parsing of RGB-D images to video sequences. In this section, we briefly summarize their method and demonstrate its extensions for the parsing of video sequences. The method takes as an input a single RGB-D view and proceeds in the following steps. The RGB image is first over segmented into superpixels which respect the straight line boundaries. RANSAC-based plane fitting estimates the dominant planes and the associated Manhattan coordinate frame of the current view, determined by one or two vertical planes perpendicular to the floor. 
The intersection of the vertical planes with the floor plane defines an infinite line; a wall may contain more than one disjointed planar segment with the same normal vector and offset. Each frame defines a bounding box volume which these lines intersect.
% Four additional walls, which are perpendicular to the floor and aligned with one of the other two axes, are added to form the boundaries of the scene. Their projection onto the floor adds four additional infinite lines. 
The end points of the lines segments together with the intersection between pairs of perpendicular infinite lines are found. These points then determine the hypothesized wall intersections and occluding boundaries. The projections of these hypothesized intersections and occluding boundaries onto an image determine the boundaries between the intervals and can be seen in Figure~\ref{fig:intervals}. The intervals then 
define the regions of the image over which the final labeling will be carried out. 
The labels are the identities of the dominant planes ${\bf l} = \{l_1, l_2, \hdots, l_k \}$, where $l_i = (n_i, d_i)$ is the plane normal and offset for one of the infinite dominant planes;  ${\bf x} = \{{x}_1, {x}_2, \hdots, {x}_k \}$ is the set of intervals, with ${\bf x_i} = (p_i, p_{i+1})$ is a segment of a field of view. We seek to assign the most likely assignment of plane labels to the set of intervals $P({\bf x} | {\bf z})$ given the depth measurements ${\bf z}$. Maximization of the probability can be rewritten as minimization of the following energy function 
\[ E(\mathbf{x}) = \sum_{i=1}^{n} (f_i (x_i, {\bf z}) + e_i(x_i, x_{i-1}, {\bf z})) \]
where $f_i(x_i)$ is the cost to assign label $l_i$ to the $i^{th}$ interval, and $e_i(x_i, x_{i-1}, {\bf z})$ is the pairwise cost of assigning to the $(i-1)^{th}$ and $i^{th}$ intervals labels $l_{i-1}$ and $l_i$, respectively. 
% Before calculating the cost to assign a label to an interval, each pixel in the frame is assigned to a best label, which is a result of a plane fitting process. When $l_i$ is considered for the $i^{th}$ interval, the endpoints of the interval are projected on to the wall $l_i$. With the ceiling and floor heights, a rectangle that defines the candidate wall segment can be found. Its projection is a quadrilateral. 
Given the estimation of the set of dominant planes definining the labels, each depth measurement is assigned the most likely plane. See Figure~\ref{fig:results} (upper left corner image) for each example. 
See Figure~\ref{fig:supports} where the cost on an interval $f_i(x_i = l_i)$ is defined as the fraction of all pixels with available depth measurements inside the quadrilateral with the best label $l_i$, divided by the total number of pixels in the quadrilateral
\[ 
c_1(x_i,l_i) = 1 - \frac{labelCount}{totalCount} \]
ranging between 0 to 1. For the labels representing virtual planes which are not supported by depth measurements, define the bounding box volume $c_1(x_i = l_i) = 0.5.$
The virtual planes not supported by any depth measurements are color coded in red in Figure~\ref{fig:results}.
The final label cost is the plane support cost weighted by the fraction of the total FOV the interval subtends
\[ f_i(x_i = l_i) = w_i . c_1(x_i, l_i). \]
The pairwise cost $e_i(x_i = l_j, x_{i-1} = l_k)$ penalizes the discontinuity when two consecutive intervals are assigned different labels. 
% In the unit circle, the two intervals have a common point; however, when they are assigned different labels, the back projection of this common point onto each wall may end up with two points that are far away from each other if the neighboring wall segments are disconnected forming an occlusion boundary.
When the optimal labeling is achieved, consecutive intervals with the same label are merged. The final result is a compact wall layout of the scene. See Figure~\ref{fig:results}. 

\subsection{Dynamic Programming for Sequences}

In a video sequence, the local structure of the scene changes very little between two consecutive frames, yet if all the frames are parsed independently, it is easy to obtain parses which are inconsistent. This is due to low quality of the depth measurements, a large amount of missing data due to reflective structures or glass, or too-oblique angles of planar structures. The brittle nature of the raw depth data further affects the process of estimating the dominant planes, determining the intervals and the labeling. Next we describe how to introduce some temporal consistency into the parsing process and obtain a locally consistent 3D layout. We will do this by incorporating the result of the previous labeling in the optimization $P({\bf x} | {\bf x'}, {\bf z})$ given the depth measurements ${\bf z}$, with ${\bf x}$ and ${\bf x'}$ denoting the set of intervals and their labeling in the current and previous frame.
% the previous extracted wall layout is used to refine the intervals, and how it is used to modify the label cost and the pairwise cost in the dynamic programming equation.

The relative pose between two consecutive frames is estimated using visual odometry, which will be discussed in Section 4.1. Given the relative pose, walls in the previous frame are associated with those in the current one. Two walls are associated if they have the same orientation, and if their offset difference is below a threshold of 0.05 m. For a plane of the previous frame that does not have any associations, it is added to the set of labels of the current frame. A list of labels is created.

In the current frame, after all the intervals have been identified using the method described in Section 3.1, a set S of interval $\{p_0, ..., p_m\}$ is found. The layout produced for the previous scene yields a collection of intervals and endpoints, which is projected to the current frame, obtaining another set S' of endpoints. Let S' be $\{p'_0, ..., p'_n\}$ and $l'_i$ is the assigned label for the $i^{th}$ interval, $[p'_{i-1}, p'_i]$, taken from the previous layout. Now, the intervals formed by a union of the end points in S and S' and a new set of labels is given by the union of the previous and current labels after plane association. Given a new set of endpoints and the intervals they induce, we now formulate the modified label costs taking into account the results of the optimal label assignment from the previous frame. 

\begin{figure}[htb]
\centering
    \includegraphics[width=0.5\textwidth]{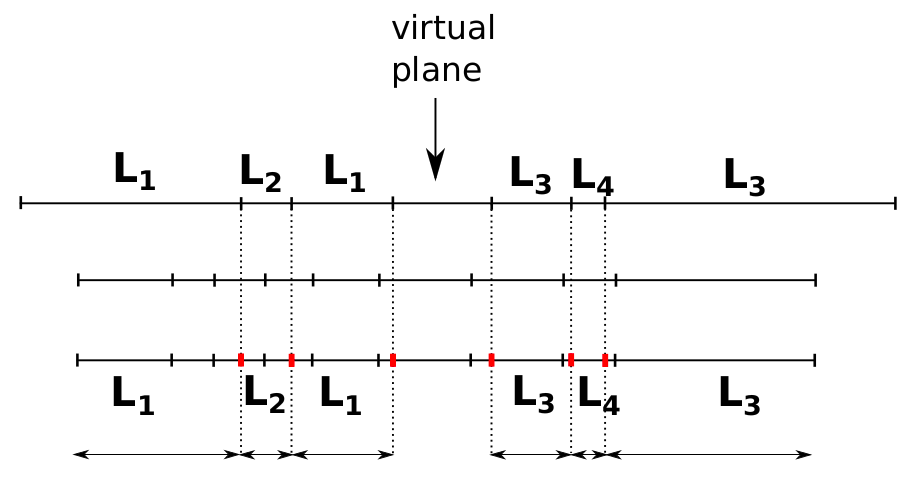}
\caption{Propagation of labels: The first horizontal line shows the endpoints of the layout from the previous frame. The second horizontal line shows the endpoints found in the current frame. The last line shows the final intervals and the propagated labels. Only the label of real planes are propagated.}
\end{figure}

Lets denote a new set of endpoints on the circle $\{s_0, ..., s_{k}\}$. When assigning label costs to a particular interval $[s_{j-1}, s_j]$, we need to consider several scenarios. 
First that there is an interval $[p'_{i-1}, p'_i]$ in the previous frame that completely covers it, with the previously assigned label $l_i$.  The cost of assigning this label again should be lower, reflecting the increased confidence in the presence of the label in the current frame, given the previous frame. 
% In single view parsing, $f_i(x_i = l_j)$ is the cost to assign label $l_j$ to the $i^{th}$ interval. When parsing a sequence, more constraints are available and can be quantified as other types of cost. In order to make it clear, let $c_1(x_i = l_j)$ be the cost used in single view parsing when label $l_j$ is assigned to the $i^{th}$ interval.
When parsing the video sequence we modify the cost function by introducing two additional costs; the fitting cost $c_2(x_i = l_j, {\bf z})$ and the temporal cost $c_3(x_i = l_j)$. The fitting cost $c_2(x_i = l_j, {\bf z})$ is the average residual for the depth measurements that lie inside the projected quadrilateral that has its best label as $l_j$. Let $\{X_1, ..., X_k\}$ be the set of these 3D points, and $l_j$ is the wall label characterized by parameters $(n_j, d_j)$.  
\begin{equation*}
	c_2(x_i = l_j, {\bf z}) = 
	\begin{cases}
		min(\frac{\sum_{i} d(X_i, l_j)}{k}, 0.15), & \text{if} \: l_j \:  \text{real wall} \\
		0.5, & \text{virtual wall}
	\end{cases}
\end{equation*}
where $d(X_i, l_j)$ is the 3D point to plane distance. This cost models the scenario where there may be more then one suitable plane model for the interval, but the plane 
fitting process has omitted the plane selection due to missing data or ambiguities. This plane label in question was however successfully detected and labeled in the previous frame and hence it is a good candidate for explaining the depth values in the interval. The suitability of the plane is measured by the average residual error. 

For the temporal cost if $l_j$ is not the preferred label, a cost of 0.1 is added, otherwise there is no penalty
\begin{equation*}
    c_3(x_i = l_j) =
    \begin{cases}
      0.1, & \text{if}\ l_j \text{is preferred} \\
      0, & \text{otherwise.}
    \end{cases}
\end{equation*}
The total cost to assign the label $l_j$ to the $i^{th}$ interval is:
\begin{equation*}
	f_i(x_i = l_j, \textbf{z}) = c_1(x_i = l_j) + c_2(x_i = l_j, {\bf z}) + c_3(x_i = l_j).
\end{equation*}
Similarly as in the single view case, the final label cost of the interval is $f_i(x_i = l_j, \textbf{z})$ is weighted by the fraction of the FOV the interval $x_i$ subtends. 

The pairwise cost is also modified to accommodate the temporal constraint. In the case that the proposed labeling introduces discontinuity of depth at the junction between the two intervals, the following penalty is applied:
\begin{equation*}
    e_i(x_i = l_j, x_{i-1} = l_k, \textbf{z}) =
    \begin{cases}
      \delta, & \text{if } \: l_j \text{ is not preferred} \\
      \frac{\delta}{3}, & \text{otherwise.}
    \end{cases}
\end{equation*}
If there is no discontinuity induced from the proposed labeling, then $e_i(x_i = l_j, x_{i-1} = l_k, \textbf{z}) = 0$.

We used a discontinuity cost of $\delta = 0.03$ in our experiments. Given that our state space ${\bf x}$ is a linear 1D chain of intervals, the optimal labeling problem can now be solved using dynamic programming as described in~\cite{TaylorRSS12}. 
% The recursive formula for temporal dynamic programming is similar to that in the case of single view, except for the extra temporal constraint.
% \begin{equation}
% D_{i,j} = \min_{k} [f_i(x_i = l_j, \textbf{z}) + e_i(x_i = % l_j, x_{i-1} = l_k, \textbf{z}) + D_{i-1,k}]
% \end{equation}

% For intervals that are covered by $[s'_{i-1}, s'_i]$, we assign a preferred label of $l_i$ for it where $l_i$ is the associated label of $l'_i$. {\bf JK Preferred labels are not assigned - you need to describe how you change the objective function. Intervals that are out of the view of the current frame are discarded. At the end of this step, a set of intervals with their preferred labels is generated. }

% On top of the label cost used in Taylor's work, we add two more. First, consider the set of pixels with a best label matching the label of the wall considered for the interval. The mean Euclidean distance between these points and the wall is calculated. Then, a cost, which is the lower of the mean distance and 0.15, is added to the label cost. For the walls that make up the boundary box, a flat cost of 0.5 is added. Second, if the considered label doesn't match the preferred label, a cost of 0.1 is added. The sum of these two new costs is weighted by the arc length formed by the two endpoints around the center of the circle normalized by the FOV. It is added to the existing label cost.

% The pairwise cost is also modified. If the $i^{th}$ interval has a preferred label of $l_i$, then the discontinuity cost $e_i(l_i, l_{i-1})$ is a third of the regular discontinuity cost.

The results of the optimal scene parsing using single view and temporal constraints is described in more detail in the experiments. 

\section{Visual Odometry}
As a result of a single view parsing we estimate the 
rotation of the camera with respect to the world coordinate 
frame $R^{cw}_i$. We omit the subscript $cw$ for clarity.  Relative rotation between consecutive frames is estimated as $R_{i-1,i} = R_{i-1}^T R_{i}$. The relative translation is estimated using SIFT matching and RANSAC requiring only a single 3D point correspondence. 
% \textit{Relative rotation estimation:} For each scene, a floor plane and a set of other planes are extracted. Using 3D points for each plane, the normal vector is estimated. Together with the normal vector of the floor plane, the relative rotation of the camera with respect to the local Manhattan frame that the wall aligns with can be estimated. The Manhattan frame that is associated with the dominant rectilinear structure is identified, and marked as the dominant Manhattan system in this frame. For other Manhattan systems, their relative rotation with respect to the dominant one is estimated. If this rotation is small, or if it is alternatively nearly perpendicular, the two system are considered to be the same.

% Then as the robot moves along a corridor, the dominant Manhattan system won't change until the robot approaches a new corridor in which the local Manhattan system doesn't align with the current one. Detecting when this change happens is important for the relative rotation estimation between two consecutive frames.

In this work we assume that only the weak Manhattan constraint is available, i.e. that the environment can have multiple local Manhattan frames. This, for example, corresponds to the settings where corridors are not orthogonal to each other~\cite{SaurerVICOMOR12}. 
% With an RGB-D sensor, the Manhattan frame can be estimated using the planes extracted from 3D point cloud. For the $i^{th}$ frame, $R_i$ is the relative rotation of the camera with respect to the local Manhattan frame.

We develop a simple but effective mechanism to detect these new frames in an online setting and adjust the process of estimation of the relative rotation $R_{i-1,i}$ accordingly.  

We assume that the first view of the sequence determines the initial word reference Manhattan frame $R_w$. In subsequent frames the single view rotation estimates are composed together to yield the rotation estimate of the camera pose with respect to the world reference frame $R_i^{cw}$. 
In the case when the single RGB-D frame has multiple vertical walls which are not perpendicular to each other, we get several estimates of the local Manhattan frame for that view, lets denote them $R_i$ and $R'_i$. 
To determine the one which will yield the correct relative rotation $R_{i-1,i} = R_{i-1}^T R_{i}$ and $R_{i-1,i} = R_{i-1}^T R'_{i}$, we choose the one which yields smaller relative rotation as the motions between consecutive frames are small. We also store the angle between $R_i$ and $R'_i$ representing the alignment between two different Manhattan frames. 

\subsection{Graph SLAM and Loop Closure Detection}
The visual odometry techniques described above yield very good rotation estimates even in the absence of features in the environment. When aligning the sequences for longer trajectories the system accumulates a small drift requiring global alignment step. We exploit the structures detected from single view reconstruction, such as walls, corners (the intersection between two walls) for the global alignment steps.

We use the commonly used global GRAPH SLAM~\cite{GraphSLAMTutorial} optimization approach. Since the height of the camera is fixed and we can estimate the single view rotation, we can always assume that camera motion is planar. In this case  the optimization is reduced to a 2D SLAM problem. The pose of the $i^{th}$ frame is denoted as \textbf{$g_i$} =  $(x_i, y_i, \theta_i)^T$. This is the pose of the camera with respect to the Manhattan coordinate frame established by the first frame.

Under the weak Manhattan assumption, the rotation of the camera can be estimated with very high accuracy without drifting, thus it's only necessary that the poses be optimized based on their locations $(x_i, y_i)^T$. Given two poses $g_i$ and $g_j$, and the observation $\hat{g}_{ij}$, the error is measured as $e_{ij}^T \Omega e_{ij}$ where $e_{ij} = g_i - g_j - \hat{g}_{ij}$ and $\Omega$ is the information matrix. The Jacobian matrices are simplified to:

\begin{equation*}
\frac{\partial {e_{ij}}}{\partial {g_i}} = \begin{bmatrix} 1 & 0\\0 & 1 \end{bmatrix} \mbox{ and } \frac{\partial {e_{ij}}}{\partial {g_j}} = \begin{bmatrix} -1 & 0\\0 & -1 \end{bmatrix} 
\end{equation*}

\textit{Building the graph:} The pose of each frame is a node in the graph. An edge is added between any two consecutive nodes and filled with information provided by the relative translation estimated using visual odometry. This relative transformation is in the global coordinate system defined when the robot just starts. The graph optimization is done every time a loop closure is detected. At the end of the sequence, one more final optimization is performed.

\begin{figure}[htb]
\centering
\includegraphics[width = 0.2\textwidth]{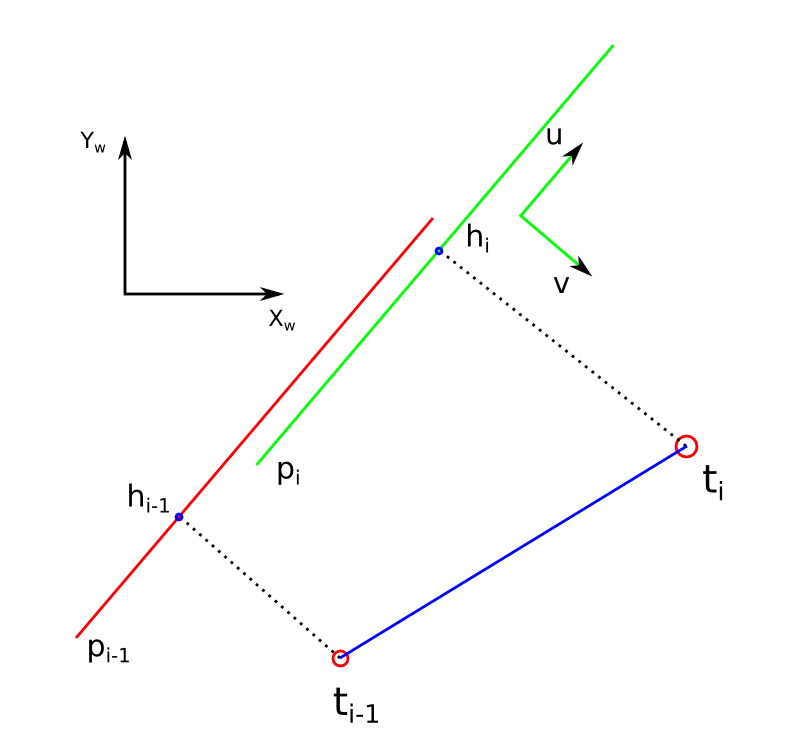}
\caption{Planar constraint for consecutive frames.}
\label{fig:planar_constraint}
\end{figure}

\textit{Pairwise constraint between consecutive frames:} Between consecutive frames, walls are associated using the joint compatibility branch and bound test~\cite{jcbb}. The dominant walls that align with the local Manhattan frame are used to enhance the consistency between two consecutive poses. As shown in Figure~\ref{fig:planar_constraint}, the estimated translation between the $(i-1)^{th}$ and $i^{th}$ frames obtained from the visual odometry is $\mathbf{t_i} - \mathbf{t_{i-1}}$. With the tracking of the planes, the enhanced constraint for the graph optimization is: $((\mathbf{t_i} - \mathbf{t_{i-1}})^{T}\mathbf{u})\mathbf{u} + ((\mathbf{t_i} - \mathbf{h_i}) - (\mathbf{t_{i-1}} - \mathbf{h_{i-1}}))$, where $\bf{v}$ is the normal vector of plane $p_i$, $\bf{u} = \bf{n} \times \bf{v}$ with $\bf{n}$ be the normal vector of the floor. $\bf h_{i-1}$ and $\bf h_i$ are the orthogonal projection of $\bf{t_{i-1}}$ and $\bf{t_i}$ on $p_{i-1}$ and $p_i$ respectively. This is used as the pairwise constraint between the $(i-1)^{th}$ and $i^{th}$ nodes when the loop closure is performed. Intuitively, this pairwise constraint enforces the co-planarity between associated walls between two consecutive frames when loop closure is run.

\begin{figure}[htb]
\centering
\includegraphics[width = 0.15\textwidth]{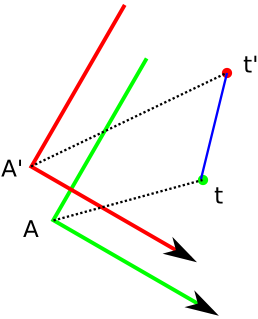}
\caption{Example of relative translation between two poses estimated at loop closure and two coordinate frames associated with corresponding corners {\bf c} and {\bf c'}.  }
\label{fig:corner_constraint}
\end{figure}

\textit{Loop Closure Detection:} The ability to recognize a place that the robot has previously visited, and then estimate the relative distance displacement of the robot between the two frames to add extra constraints to the graph is important to get a good pose optimization. In our work, the loop closure detection is done by using GIST features~\cite{Oliva2001} at places marked as geometric signatures of the sequence. A frame is marked as a key frame if it contains at least one geometric signature, a pair of walls that are orthogonal to each other and connected. In general, geometric signatures are found at intersections, T-junctions, and corner turns. When such frame is detected, a matching process will be carried out to find out if it matches with any previously found key frames. For each match, an edge is added to the pose graph. The criteria for the matching is: the relative rotation between the two pairs is less than $10^o$, the distance between their locations is less than 5 meters, and finally the GIST score between the two scenes is less than 0.025. In case there's a match, the relative displacement between the two frames is estimated using the matching pairs of orthogonal and connected walls as: 
\begin{eqnarray}
\bf (c - t) - (c' - t')
\nonumber
\end{eqnarray}
where ${\bf c} = [x_c, y_c]^T$ and ${\bf c}'=[x'_c, y'_c]^T$ are the matched corners. See Figure~\ref{fig:corner_constraint}.
\subsection{Final Global Map Generation} 

When the globally refined poses are found, the locations of walls in each frame are updated. At the end of the sequence, walls are merged frame after frame to generate the global wall maps.

\textit{Generate the coarse map:} First, big walls (those with a length of at least 2 meters) are merged to generate a coarse map. In Figures~\ref{fig:map1} and~\ref{fig:map2}, the coarse map consists of the red lines that are 1 meter or longer. The criteria for merging are: the angle between two walls is less than $5^o$, distance between two walls, which is measured by the maximum distance between each endpoint to the other wall, is less than 0.25 meters, the sum of the absolute difference of the average color of the walls in three channels (hue, sat, value) is less than 30.

\begin{figure}[htb]
\centering
\includegraphics[width=0.15\textwidth]{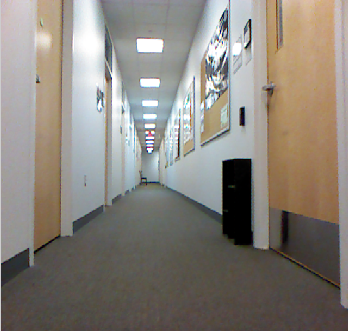}
\hspace{0.05cm}
\includegraphics[width=0.15\textwidth]{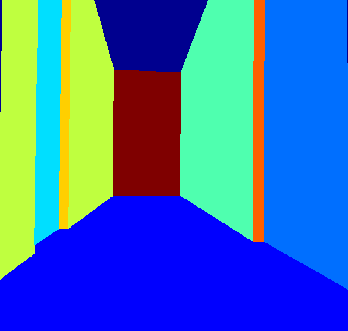} \\
\vspace{0.2cm}
\includegraphics[width=0.15\textwidth]{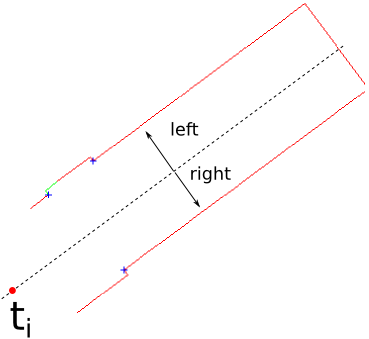}
\hspace{0.05cm}
\includegraphics[width=0.15\textwidth]{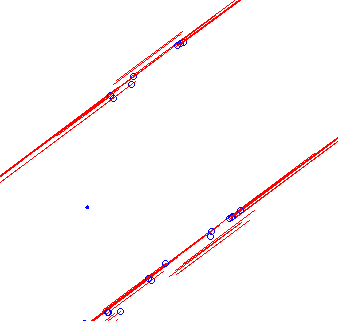}
\caption{Top: RGB image and the corresponding single view reconstruction; Bottom: The left is the projection of the walls onto the ground floor, the blue marks are the locations of detected corners. The right is a superposition of walls (red lines) and corners (blue circles) of several frames.}
\label{fig:door_corner}
\end{figure}

\textit{Door detection:} To detect doors, we keep track of a set of corners detected in all the frames. These corners are the ends of the innermost walls in the left and the right of the camera, as shown in the bottom left part of Figure~\ref{fig:door_corner}, denoted with blue cross marks. A wall is a door candidate if it is not wider than 1 meter (we assume that door has a width of about 0.825 meters) and has at least 2 corners near each end (within 0.25 meters). Door candidates will be merged if the angle between them is less than $5^o$, the distance between two walls is less than 0.25 meters, and they overlap with the intersection over union score by at least 0.25. Once doors are extracted, they are added to the map. For the small walls that do not pass the door test, we merge them with the big walls in the coarse map.

\section{Experiments}

We evaluate our algorithm on several RGB-D sequences of indoor scenes with minimal texture, that satisfy Manhattan or weak Manhattan constraints. One of the sequences is from the TUM RGB-D dataset~\cite{sturm11rss-rgbd}, \textit{fr3/structure-notexture-far}, which is publicly available and comes with ground truth. Besides this, we collected two other sequences of large scale office corridors and tested them. 

\subsection{Temporal Parsing}

Our experiments demonstrate that temporal parsing produced incrementally better results in scenarios where depth data was missing or noisy due to sensor limitations, or due to the difficult nature of data such as glass doors and glass walls. Our algorithm also consistently detected door planes once they had been picked up. 

Qualitative results of temporal parsing for five different scenarios, each consisting of three consecutive frames, are shown in Figure~\ref{fig:results}. For each frame, the top row shows the RGB image on the left, and walls aligning with the dominant Manhattan frame on the right. The bottom row shows the result of single view parsing on the left, and that of temporal parsing on the right. The first frame of each scenario is the starting frame, so the results for the single view and temporal parsing are the same. Scenario 1 demonstrates that temporal parsing consistently picked up doors while single view parsing failed. Scenarios 2 and 5 shows that temporal parsing produced better results for small walls in complex scene. An enclave area was correctly parsed in Scenario 3. In Scenario 4, after picking up the first frame, temporal parsing could infer a plane for the glass area in consecutive frames, while single view parsing assigned a virtual plane for it.

\subsection{Graph SLAM for weak Manhattan Indoor Environments}

\begin{figure}[h]
\noindent
\centering
\includegraphics[width=0.25\textwidth]{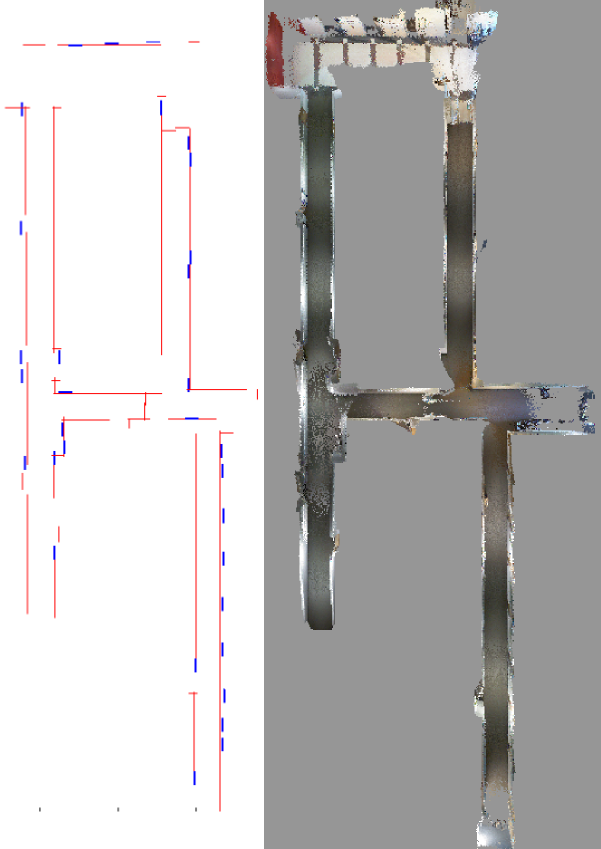}
\caption{Global maps and reconstructed point cloud of the first sequence (scales are not the same).}
\label{fig:map1}
\end{figure}

\begin{figure}[h]
\noindent
\centering
\includegraphics[width=0.4\textwidth]{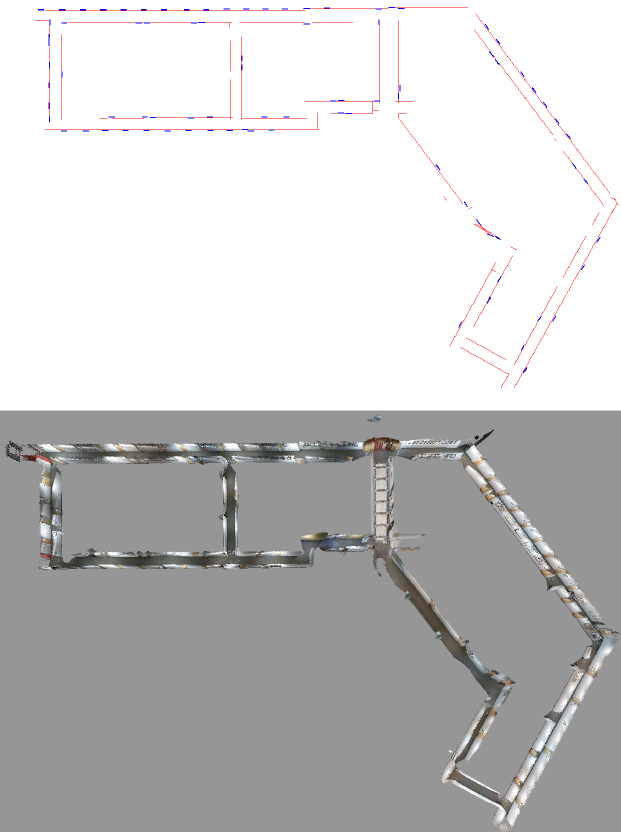}
\caption{Global maps and reconstructed point cloud of the second sequence (scales are not the same).}
\label{fig:map2}
\end{figure}

We show next that under the weak Manhattan assumption, graph SLAM optmization could be carried out on the positions of the cameras only as the estimated rotations were good and drift free. For each test sequence, a global map was generated. See Figures~\ref{fig:map1} and~\ref{fig:map2}. For comparison we selected DVO-RGBD SLAM~\cite{dvo_slam} and ORB-RGBD SLAM~\cite{orb_rgbd_slam} to demonstrate inferior results compared to our algorithm.

\textit{TUM SLAM dataset:} Our algorithm focuses on SLAM for scenes with Manhattan/weak Manhattan structure without features. The sequence \textit{fr3/structure-notexture-far} is a top candidate to demonstrate our approach as it meets most of these constraints. Besides, it comes with the ground truth that allows a comparison between our algorithm and other methods. 

\begin{figure}[htb]
\centering
\includegraphics[width = 0.3\textwidth]{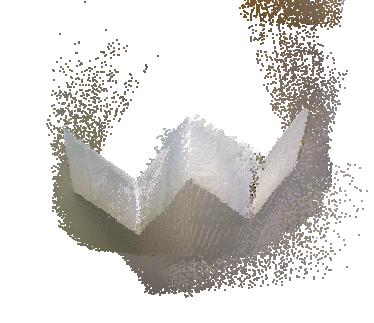}
\caption{Reconstructed point cloud using estimated poses.}
\label{fig:pl_snf}
\end{figure}

\textit{Qualitative Result:} A densely reconstructed point cloud using the trajectory generated from our slam framework is shown in Figure ~\ref{fig:pl_snf}.

\begin{figure}[htb]
\centering
\includegraphics[width = 0.32\textwidth]{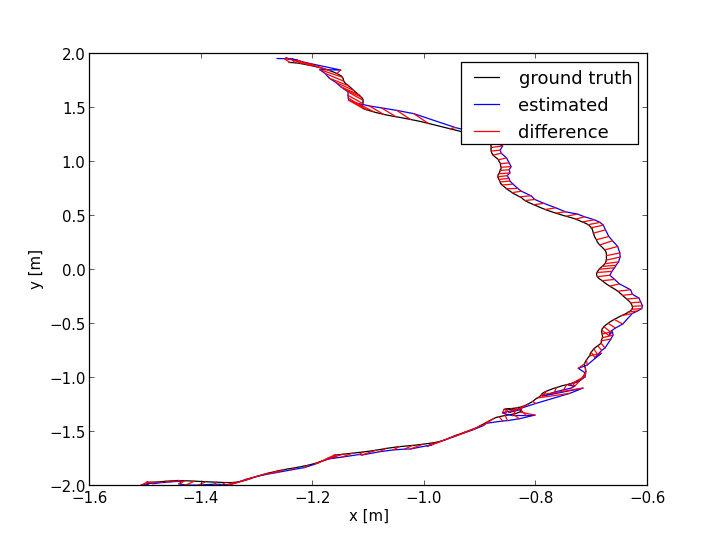}
\caption{Our estimated trajectory versus the provided ground truth.}
\label{fig:gt_vs_our}
\end{figure}

\textit{Quantitative Result:} With the availability of the ground truth, the root mean squared error (RMSE) can be computed. We ran our algorithm five times on the same sequence to obtain the average RMSE and the deviation. We also followed the same procedure for DVO-RGBD SLAM and ORB-RGBD SLAM. The comparison is shown in Table ~\ref{tab:1}. Besides DVO-RGBD and ORB-RGBD, we also include the result of Pop-up Plane Slam (taken directly from their paper~\cite{popupslam}, which is not a RGB-D SLAM framework but it is relevant. The comparison is shown in Table~\ref{tab:1}. The difference between the trajectory generated by our algorithm and the provided ground truth is shown in Figure ~\ref{fig:gt_vs_our}.

%- Pop-up Plane SLAM generates a pop-up model for each single RGB image. The model consists of a floor plane and walls perpendicular to it. The floor plane and walls are then fed to a planar SLAM framework. \\
%
%- DVO slam: DVO slam is currently one of the best rgb-d slam algorithm. The algorithm makes use of both photometric and depth error over the pixels, together with loop closure.\\
%
%- ORB-RGBD slam:\\

\begin{table}
\centering
\caption{Results for TUM RGB-D Dataset}
\label{tab:1}
\begin{tabular}{ |c|c| } 
 \hline
  & RMSE (m) \\ 
  \hline
 Pop-up Plane SLAM & $0.18 \pm 0.07$ \\ 
 \hline
 DVO-RGBD SLAM & $0.097 \pm 0.000$ \\ 
 \hline
 ORB-RGBD SLAM & $0.016 \pm 0.002$ \\
 \hline
 Ours & $0.043 \pm 0.001$\\
 \hline
\end{tabular} 
\end{table}

\begin{figure}[htb]
\centering
\includegraphics[width=0.22\textwidth]{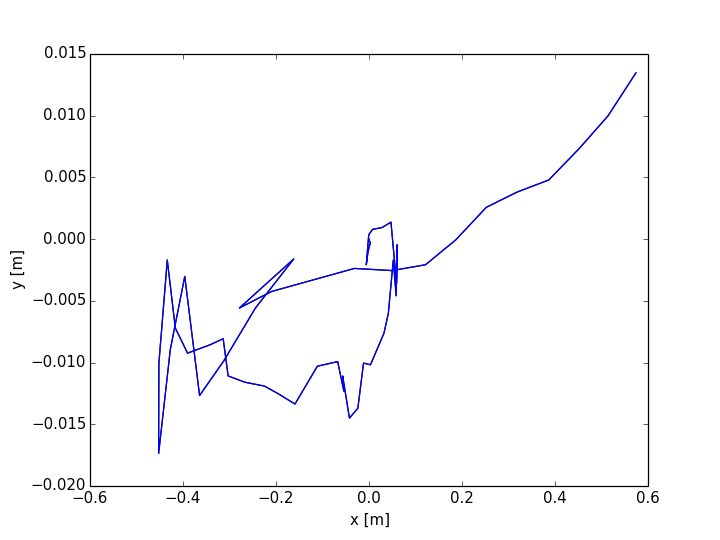}
\includegraphics[width=0.22\textwidth]{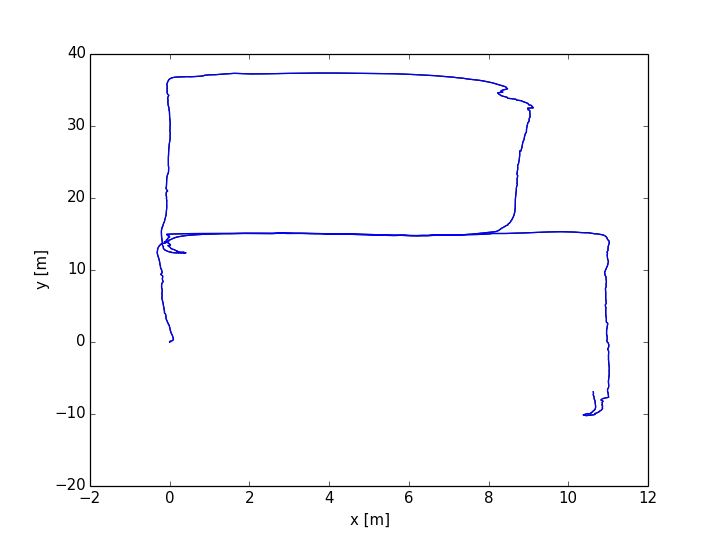}\\
\includegraphics[width=0.22\textwidth]{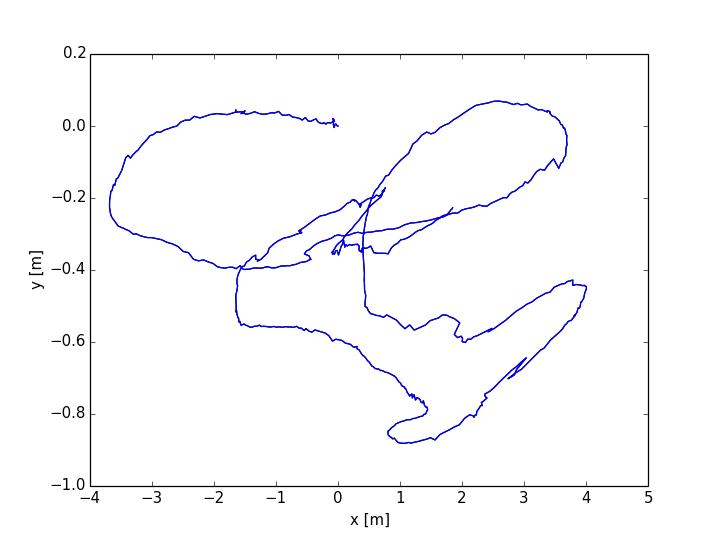}
\includegraphics[width=0.22\textwidth]{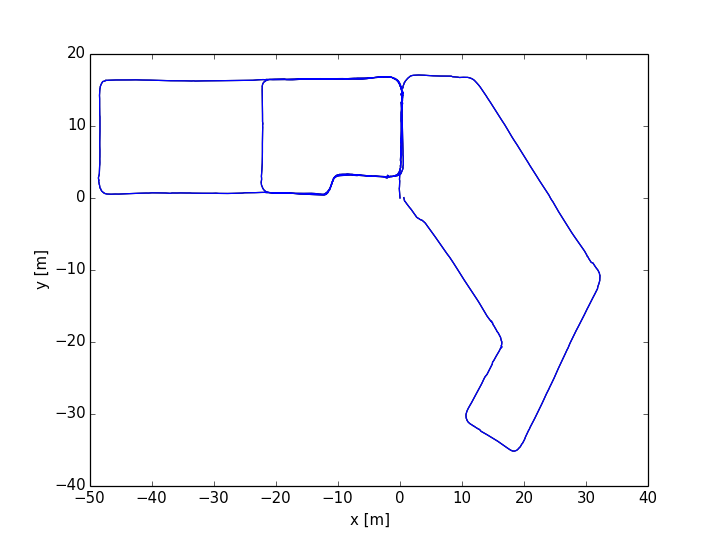}
\caption{Left column: DVO SLAM trajectory for the sequences. Right column: our results.}
\label{fig:slam_1}
\end{figure}

\textit{Large Scale Indoor Office:} We do not have the ground truth for these sequences, thus only qualitative comparison is possible. As the source code for Pop-up Plane SLAM is not available, we only compare our algorithm with DVO-RGBD SLAM and ORB-RGBD SLAM. DVO-RGBD SLAM produced a meaningless trajectory for both sequences, while ORB-RGBD SLAM kept losing the tracking and did not produce a complete trajectory for the sequences. Figure~\ref{fig:slam_1} shows the result of our algorithm versus DVO-RGBD SLAM. Figures~\ref{fig:map1} and~\ref{fig:map2} show the point cloud reconstruction of the two sequences using our estimated poses.

\textit{Door detection:} A summary for the door detection is shown in Table~\ref{tab:2}.

\begin{table}
\centering
\caption{}
\label{tab:2}
\begin{tabular}{ |c|c|c|c| } 
 \hline
  seq no & doors detected & correct detection & missed\\ 		
  \hline
  1 & 32 & 18 & 7\\ 
  \hline
  2 & 86 & 77 & 7\\
  \hline
\end{tabular} 
\end{table}

It is noticeable that DVO SLAM and ORB-RGBD SLAM work well for the TUM sequence, which does not have texture, but fails on our sequences. There is a major difference between the TUM sequence and ours. Even though the TUM sequence is textureless, many reliable point features can still be detected and tracked. For our indoor office sequences, the scene often consists just of blank walls and a few distinct feature points. For our case, as rotation is reliably estimated from the structure of the scene, only a few matching point features are needed to estimate the translation, which is not the case for DVO-SLAM and ORB-RGBD SLAM.

\section{Conclusions}

We have presented a temporal parsing algorithm that yields better, temporally consistent results in challenging RGB-D sequences. The algorithm consistently and correctly parses meaningful structures in the scene such as door planes. This enables an efficient on-line method to detect doors which were not propagated to the final global map. We have also introduced an efficient visual odometry algorithm that works without rotation drift in a weak Manhattan world setting. Finally, pose optimization based on the locations of the camera and the constraints obtained by matching geometric signatures between key frames provides global refinement of poses. At the end of the pipeline, a global map for the sequence is generated.

% \begin{figure}[h]
% \noindent
% \centering
% \includegraphics[width=0.3\textwidth]{hw1_1.png}
% \caption{The reconstructed point clouds using our estimated poses for the first sequence.}
% \label{fig:map4}
% \end{figure}

% \begin{figure*}[h]
% \noindent
% \centering
% \includegraphics[width=0.9\textwidth]{final_maps_hw3_1.png}
% \caption{Global maps of a sequence where the weak Manhattan constraint is satisfied.}
% \label{fig:map2}
% \end{figure*}

{\small
\bibliographystyle{ieee}
\bibliography{bibfile}
}

\end{document}